# Marginalized Denoising Autoencoders for Domain Adaptation

**Minmin Chen**  MC15@CSE.WUSTL.EDU
**Zhixiang (Eddie) Xu**  XUZX@CSE.WUSTL.EDU
**Kilian Q. Weinberger**  KILIAN@WUSTL.EDU
Washington University, St. Louis, MO 63130, USA

**Fei Sha**  FEISHA@USC.EDU
U. of Southern California, Los Angeles, CA 90089, USA

## Abstract

Stacked denoising autoencoders (SDAs) have been successfully used to learn new representations for domain adaptation. Recently, they have attained record accuracy on standard benchmark tasks of sentiment analysis across different text domains. SDAs learn robust data representations by reconstruction, recovering original features from data that are artificially corrupted with noise. In this paper, we propose *marginalized SDA* (mSDA) that addresses two crucial limitations of SDAs: high computational cost and lack of scalability to high-dimensional features. In contrast to SDAs, our approach of mSDA marginalizes noise and thus does not require stochastic gradient descent or other optimization algorithms to learn parameters — in fact, they are computed in closed-form. Consequently, mSDA, which can be implemented in only 20 lines of MATLAB$^{\text{TM}}$, significantly speeds up SDAs by two orders of magnitude. Furthermore, the representations learnt by mSDA are as effective as the traditional SDAs, attaining almost identical accuracies in benchmark tasks.

## 1. Introduction

Domain adaptation (Ben-David et al., 2009; Huang et al., 2007; Weinberger et al., 2009; Xue et al., 2008) aims to generalize a classifier that is trained on a *source* domain, for which typically plenty of training data is available, to a *target* domain, for which data is scarce. Cross-domain generalization is important in many application areas of machine learning, where such an imbalance of training data may oc-cur. Examples are computational biology (Liu et al., 2008), natural language processing (Daume III, 2007; McClosky et al., 2006) and computer vision (Saenko et al., 2010).

Data in the *source* and the *target* are often distributed differently. This presents a major obstacle in adapting predictive models. Recent work has investigated several techniques for alleviating the difference: instance reweighting (Huang et al., 2007; Mansour et al., 2009), sub-sampling from both domains (Chen et al., 2011b) and learning joint target and source feature representations (Blitzer et al., 2006; Glorot et al., 2011; Xue et al., 2008).

Recently, Glorot et al. (2011) proposed a new approach that falls into the third category. The authors propose to learn robust feature representations with stacked denoising autoencoders (SDA) (Vincent et al., 2008). Denoising autoencoders are one-layer neural networks that are optimized to reconstruct input data from partial and random corruption. These denoisers can be stacked into deep learning architectures. The outputs of their intermediate layers are then used as input features for SVMs (Lee et al., 2009). Glorot et al. (2011) demonstrate that using SDA-learned features in conjunction with linear SVM classifiers yields record performance on the benchmark tasks of sentiment analysis across different product domains (Blitzer et al., 2006).

Despite their remarkable and promising results, SDAs are limited by their high computational cost. They are significantly slower to train than competing algorithms (Blitzer et al., 2006; Chen et al., 2011a; Xue et al., 2008), primarily because of their reliance on iterative and numerical optimization to learn model parameters. The challenge is further compounded by the dimensionality of the input data and the need for computationally intensive model selection procedures to tune hyperparameters. Consequently, even a highly optimized implementation (Bergstra et al., 2010) may require hours (even days) of training time.

In this paper, we address this challenge with a variant of SDA. The proposed method, which we refer to as *marginalized Stacked Denoising Autoencoder* (mSDA), adopts the greedy layer-by-layer training of SDAs. However, a crucial difference is that we use *linear* denoisers as the basic building blocks. The key observation is that, in this setting, the random feature corruption can be marginalized out. Conceptually, this is equivalent to training the models with an infinitely large number of corrupted input data. Fitting models on such a scale would be impossible for the conventional SDAs, which often rely on stochastic gradient descent, and need to sweep through all the training data.

Our contributions are summarized as follows: i) we contribute to deep learning by demonstrating that linear denoisers can be used as building blocks for learning feature representations. ii) we show that linearity can significantly simplify parameter estimation — our approach results in closed-form solutions for the optimal parameters. iii) we evaluate our approach rigorously on established domain adaptation benchmark data sets and compare with several competing state-of-the-art algorithms. We show that the classification performance of mSDA matches that of SDA across our benchmark data sets, while achieving tremendous speedups during training time (reducing training from up to 2 days for SDA to a few minutes with mSDA).

## 2. Notation and Background

We follow the setup of Glorot et al. (2011) and focus on the problem of domain adaptation throughout this paper. We assume that our data originates from two domains, source $S$ and target $T$. From the source domain $S$, we sample data $D_S = \{\mathbf{x}_1, \ldots, \mathbf{x}_{n_s}\} \subset \mathcal{R}^d$ with *known labels* $L_S = \{y_1, \ldots, y_{n_s}\}$, whereas from the target domain we are only able to sample data *without labels* $D_T = \{\mathbf{x}_{n_s+1}, \ldots \mathbf{x}_n\} \subset \mathcal{R}^d$. We *do not* assume that both domains use identical features and we pad all input vectors with zeros to make both domains be of equal dimensionality $d$. Our goal is to learn a classifier $h \in \mathcal{H}$ with the help of the labeled set $D_S$ and the unlabeled set $D_T$, to accurately predict the labels of data from the target domain $T$. In practice (and as we show in section 5) it is straightforward to also extend this framework to multiple target domains.

**Stacked Denoising Autoencoder.** Various forms of autoencoders have been developed in the deep learning literature (Rumelhart et al., 1986; Baldi & Hornik, 1989; Kavukcuoglu et al., 2009; Lee et al., 2009; Vincent et al., 2008; Rifai et al., 2011). In its simplest form, an autoencoder has two components, an encoder $h(\cdot)$ maps an input $\mathbf{x} \in \mathcal{R}^d$ to some hidden representation $h(\mathbf{x}) \in \mathcal{R}^{d_h}$, and a decoder $g(\cdot)$ maps this hidden representation back to a reconstructed version of $\mathbf{x}$, such that $g(h(\mathbf{x})) \approx \mathbf{x}$. The parameters of the autoencoders are learned to minimize the reconstruction error, measured by some loss $\ell(\mathbf{x}, g(h(\mathbf{x})))$.

Choices for the loss include squared error or Kullback-Leibler divergence when the feature values are in $[0, 1]$.

Denoising Autoencoders (DAs) incorporate a slight modification to this setup and corrupt the inputs before mapping them into the hidden representation. They are trained to reconstruct (or *denoise*) the original input $\mathbf{x}$ from its corrupted version $\tilde{\mathbf{x}}$ by minimizing $\ell(\mathbf{x}, g(h(\tilde{\mathbf{x}})))$. Typical choices of corruption include additive isotropic Gaussian noise or binary masking noise. In this work, as in Vincent et al. (2008), we use the latter and set a fraction of the features of each input to *zero*. This is a natural choice for bag-of-word representations of texts, where typical class-specific words can be missing due to the writing style of the author or differences between train and test domains.

The stacked denoising autoencoder (SDA) of Vincent et al. (2008) stacks several DAs together to create higher-level representations, by feeding the hidden representation of the $t^{th}$ DA as input into the $(t+1)^{th}$ DA. The training is performed greedily, layer by layer.

**Feature Generation.** Many researchers have seen autoencoders as a powerful tool for automatic discovery and extraction of nonlinear features. For example, Lee et al. (2009) demonstrate that the hidden representations computed by either all or partial layers of a convolutional neural network (CNN) make excellent features for classification with SVMs. The pre-processing with a CNN improves the generalization by increasing robustness against noise and label-invariant transformations.

Glorot et al. (2011) successfully apply SDAs to extract features for domain adaptation in document sentiment analysis. The authors train an SDA to reconstruct the input vectors (ignoring the labels) on the union of the source and target data. A classifier (*e.g.* a linear SVM) trained on the resulting feature representation $h(\mathbf{x})$ transfers significantly better from source to target than one trained on $\mathbf{x}$ directly. Similar to CNNs, SDAs also combine correlated input dimensions, as they reconstruct removed feature values from uncorrupted features. It is shown that SDAs are able to disentangle hidden factors which explain the variations in the input data, and automatically group features in accordance with their relatedness to these factors (Glorot et al., 2011). This helps transfer across domains as these generic concepts are invariant to domain-specific vocabularies.

As an intuitive example, imagine that we classify product reviews according to their sentiments. The source data consists of *book* reviews, the target of *kitchen appliances*. A classifier trained on the original source never encounters the bigram "energy efficient" during training and therefore assigns zero weight to it. In the learned SDA representation, the bigram "energy efficient" would tend to reconstruct, and be reconstructed by, co-occurring features, typ-ically of similar sentiment (*e.g.* "good" or "love"). Hence, the source-trained classifier can assign weights even to features that never occur in its original domain representation, which are "re-constructed" by the SDA.

Although SDAs generate excellent features for domain adaptation, they have several drawbacks: 1. Training with (stochastic) gradient descent is slow and hard to parallelize (although a dense-matrix GPU implementation exists (Bergstra et al., 2010) and an implementation based on reconstruction sampling exists (Dauphin Y., 2011) for sparse inputs); 2. There are several hyper-parameters (learning rate, number of epochs, noise ratio, mini-batch size and network structure), which need to be set by cross validation — this is particularly expensive as each individual run can take several hours; 3. The optimization is inherently non-convex and dependent on its initialization.

## 3. SDA with Marginalized Corruption

In this section we introduce a modified version of SDA, which preserves its strong feature learning capabilities, and alleviates the concerns mentioned above through speedups of several orders of magnitudes, fewer meta-parameters, faster model-selection and layer-wise convexity.

### 3.1. Single-layer Denoiser

The basic building block of our framework is a one-layer denoising autoencoder. We take the inputs $\mathbf{x}_1, \ldots, \mathbf{x}_n$ from $D = D_S \cup D_T$ and corrupt them by random feature removal — each feature is set to 0 with probability $p \geq 0$. Let us denote the corrupted version of $\mathbf{x}_i$ as $\tilde{\mathbf{x}}_i$. As opposed to the two-level *encoder* and *decoder* in SDA, we reconstruct the corrupted inputs with a single mapping $\mathbf{W} : \mathcal{R}^d \to \mathcal{R}^d$, that minimizes the squared reconstruction loss

$$\frac{1}{2n} \sum_{i=1}^{n} \|\mathbf{x}_i - \mathbf{W}\tilde{\mathbf{x}}_i\|^2. \quad (1)$$

To simplify notation, we assume that a constant feature is added to the input, $\mathbf{x}_i = [\mathbf{x}_i; 1]$, and an appropriate bias is incorporated within the mapping $\mathbf{W} = [\mathbf{W}, \mathbf{b}]$. The constant feature is *never* corrupted.

The solution to (1) depends on which features of each input are randomly corrupted. To lower the variance, we perform multiple passes over the training set, each time with different corruption. We solve for the $\mathbf{W}$ that minimizes the overall squared loss

$$\mathcal{L}_{sq}(\mathbf{W}) = \frac{1}{2mn} \sum_{j=1}^{m} \sum_{i=1}^{n} \|\mathbf{x}_i - \mathbf{W}\tilde{\mathbf{x}}_{i,j}\|^2, \quad (2)$$

where $\tilde{\mathbf{x}}_{i,j}$ represents the $j^{th}$ corrupted version of the original input $\mathbf{x}_i$.

**Algorithm 1** mDA in MATLAB$^{\text{TM}}$.

```
function [W,h]=mDA(X,p);
X=[X;ones(1,size(X,2))];
d=size(X,1);
q=[ones(d-1,1).*(1-p); 1];
S=X*X';
Q=S.*(q*q');
Q(1:d+1:end)=q.*diag(S);
P=S.*repmat(q',d,1);
W=P(1:end-1,:)/(Q+1e-5*eye(d));
h=tanh(W*X);
```

Let us define the design matrix $\mathbf{X} = [\mathbf{x}_1, \ldots, \mathbf{x}_n] \in \mathcal{R}^{d \times n}$ and its $m$-times repeated version as $\overline{\mathbf{X}} = [\mathbf{X}, \ldots, \mathbf{X}]$. Further, we denote the corrupted version of $\overline{\mathbf{X}}$ as $\widetilde{\mathbf{X}}$. With this notation, the loss in eq. (1) reduces to

$$\mathcal{L}_{sq}(\mathbf{W}) = \frac{1}{2nm}\text{tr}\left[\left(\overline{\mathbf{X}} - \mathbf{W}\widetilde{\mathbf{X}}\right)^\top \left(\overline{\mathbf{X}} - \mathbf{W}\widetilde{\mathbf{X}}\right)\right]. \quad (3)$$

The solution to (3) can be expressed as the well-known closed-form solution for ordinary least squares (Bishop, 2006):

$$\mathbf{W} = \mathbf{P}\mathbf{Q}^{-1} \text{ with } \mathbf{Q} = \widetilde{\mathbf{X}}\widetilde{\mathbf{X}}^\top \text{ and } \mathbf{P} = \overline{\mathbf{X}}\widetilde{\mathbf{X}}^\top. \quad (4)$$

(In practice this can be computed as a system of linear equations, without the costly matrix inversion.)

### 3.2. Marginalized Denoising Autoencoder

The larger $m$ is, the more corruptions we average over. Ideally we would like $m \to \infty$, effectively using infinitely many copies of noisy data to compute the denoising transformation $\mathbf{W}$.

By the weak law of large numbers, the matrices $\mathbf{P}$ and $\mathbf{Q}$, as defined in (3), converge to their expected values as $m$ becomes very large. If we are interested in the limit case, where $m \to \infty$, we can derive the expectations of $\mathbf{Q}$ and $\mathbf{P}$, and express the corresponding mapping $\mathbf{W}$ as

$$\mathbf{W} = E[\mathbf{P}]E[\mathbf{Q}]^{-1}. \quad (5)$$

In the remainder of this section, we compute the expectations of these two matrices. For now, let us focus on

$$E[\mathbf{Q}] = \sum_{i=1}^{n} E\left[\tilde{\mathbf{x}}_i \tilde{\mathbf{x}}_i^\top\right]. \quad (6)$$

An off-diagonal entry in the matrix $\tilde{\mathbf{x}}_i \tilde{\mathbf{x}}_i^\top$ is uncorrupted if the two features $\alpha$ and $\beta$ both "survived" the corruption, which happens with probability $(1-p)^2$. For the diagonal entries, this holds with probability $1-p$. Let us define a vector $\mathbf{q} = [1-p, \ldots, 1-p, 1]^\top \in \mathcal{R}^{d+1}$, where $\mathbf{q}_\alpha$ represents the probability of a feature $\alpha$ "surviving" the corruption. As the constant feature is never corrupted, we have $\mathbf{q}_{d+1} = 1$. If we further define the scatter matrix of the original uncorrupted input as $\mathbf{S} = \mathbf{X}\mathbf{X}^\top$, we can express the expectation of the matrix $Q$ as

$$E[\mathbf{Q}]_{\alpha,\beta} = \begin{cases} \mathbf{S}_{\alpha\beta}\mathbf{q}_\alpha\mathbf{q}_\beta & \text{if } \alpha \neq \beta \\ \mathbf{S}_{\alpha\beta}\mathbf{q}_\alpha & \text{if } \alpha = \beta \end{cases}. \quad (7)$$

Similarly, we obtain the expectation of $\mathbf{P}$ in closed-form as $E[\mathbf{P}]_{\alpha\beta} = \mathbf{S}_{\alpha\beta}\mathbf{q}_\beta$.

With the help of these expected matrices, we can compute the reconstructive mapping $\mathbf{W}$ directly in closed-form without ever explicitly constructing a single corrupted input $\tilde{\mathbf{x}}_i$. We refer to this algorithm as marginalized Denoising Autoencoder (mDA). Algorithm 1 shows a 10-line MATLAB™ implementation. The mDA has several advantages over traditional denoisers: 1. It requires only a single sweep through the data to compute the matrices $E[\mathbf{Q}], E[\mathbf{P}]$; 2. Training is convex and a globally optimal solution is guaranteed; 3. The optimization is performed in non-iterative closed-form.

### 3.3. Nonlinear feature generation and stacking

Arguably two of the key contributors to the success of the SDA are its *nonlinearity* and the *stacking* of multiple layers of denoising autoencoders to create a "deep" learning architecture. Our framework has the same capabilities.

In SDAs, the nonlinearity is injected through the nonlinear *encoder* function $h(\cdot)$, which is learned together with the reconstruction weights $\mathbf{W}$. Such an approach makes the training procedure highly non-convex and requires iterative procedures to learn the model parameters. To preserve the closed-form solution from the linear mapping in section 3.2 we insert nonlinearity into our learned representation *after* the weights $\mathbf{W}$ are computed. A nonlinear squashing-function is applied on the output of each mDA. Several choices are possible, including sigmoid, hyperbolic tangent, $\tanh()$, or the rectifier function (Nair & Hinton, 2010). Throughout this work, we use the $\tanh()$ function.

Inspired by the layer-wise stacking of SDA, we stack several mDA layers by feeding the output of the $(t-1)^{th}$ mDA (after the squashing function) as the input into the $t^{th}$ mDA. Let us denote the output of the $t^{th}$ mDA as $\mathbf{h}^t$ and the original input as $\mathbf{h}^0 = \mathbf{x}$. The training is performed greedily layer by layer: each map $\mathbf{W}^t$ is learned (in closed-form) to reconstruct the previous mDA output $\mathbf{h}^{t-1}$ from all possible corruptions and the output of the $t^{th}$ layer becomes $\mathbf{h}^t = \tanh(\mathbf{W}^t \mathbf{h}^{t-1})$. In our experiments, we found that even without the nonlinear squashing function, stacking still improves the performance. However, the nonlinearity improves over the linear stacking significantly. We refer to the stacked denoising algorithm as marginalized Stacked Denoising Autoencoder (mSDA). Algorithm 2 shows a 8-lines MATLAB™ implementation of mSDA.

**Algorithm 2** mSDA in MATLAB™.
```
function [Ws,hs]=mSDA(X,p,l);
 [d,n]=size(X);
 Ws=zeros(d,d+1,l);
 hs=zeros(d,n,l+1);
 hs(:,:,1)=X;
 for t=1:l
  [Ws(:,:,t), hs(:,:,t+1)]=mDA(hs(:,:,t),p);
 end;
```

### 3.4. mSDA for Domain Adaptation

We apply mSDA to domain adaptation by first learning features in an unsupervised fashion on the union of the source and target data sets. One observation reported in (Glorot et al., 2011) is that if multiple domains are available, sharing the unsupervised pre-training of SDA across all domains is beneficial compared to pre-training on the source and target only. We observe a similar trend with our approach. The results reported in section 5 are based on features learned on data from all available domains. Once a mSDA is trained, the output of all layers, after squashing, $\tanh(\mathbf{W}^t \mathbf{h}^{t-1})$, combined with the original features $\mathbf{h}^0$, are concatenated and form the new representation. All inputs are transformed into the new feature space. A linear Support Vector Machine (SVM) (Chang & Lin, 2011) is then trained on the transformed source inputs and tested on the target domain. There are two meta-parameters in mSDA: the corruption probability $p$ and the number of layers $l$. In our experiments, both are set with 5-fold cross validation on the labeled data from the *source* domain. As the mSDA training is almost instantaneous, this grid search is almost entirely dominated by the SVM training time.

## 4. Extension for High Dimensional Data

Many data sets (*e.g.* bag-of-words text documents) are naturally high dimensional. As the dimensionality increases, hill-climbing approaches used in SDAs can become prohibitively expensive. In practice, a work-around is to truncate the input data to the $r \ll d$ most common features (Glorot et al., 2011). Unfortunately, this prevents SDAs from utilizing important information found in rarer features. (As we show in section 5, including these rarer features leads to significantly better results.) High dimensionality also poses a challenge to mSDA, as the system of linear equations in (5) of complexity $O(d^3)$ becomes too costly. In this section we describe how to approximate this calculation with a simple division into $\frac{d}{r}$ sub-problems of $O(r^3)$.

We combine the concept of "pivot features" from Blitzer et al. (2006) and the use of most-frequent features from Glorot et al. (2011). Instead of learning a single mapping $\mathbf{W} \in \mathcal{R}^{d \times (d+1)}$ to reconstruct all corrupted features, we learn *multiple mappings* but only reconstruct the $r \ll d$ most frequent features (here, $r = 5000$). For an input $\mathbf{x}_i$ we denote the shortened $r$-dimensional vector of only the $r$ most-frequent features as $\mathbf{z}_i \in \mathcal{R}^r$. We perform this reconstruction with $S$ random non-overlapping sub-sets of input features. Without loss of generality, we assume that the feature-dimensions in the input space are in random order and divide-up the input vectors as $\mathbf{x}_i = \left[\mathbf{x}_i^{1\top}, \ldots, \mathbf{x}_i^{S\top}\right]^\top$. For each one of these sub-spaces we learn an independent mapping $\mathbf{W}^s$ which minimizes

$$\mathcal{L}_s(\mathbf{W}^s) = \frac{1}{2n}\sum_{i=1}^n \sum_{s=1}^S \|\mathbf{z}_i - \mathbf{W}^s \tilde{\mathbf{x}}_i^s\|^2. \quad (8)$$

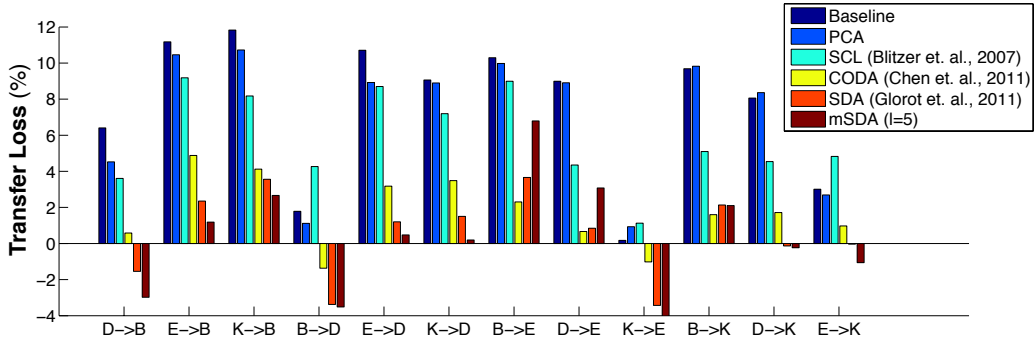

*Figure 1.* Detailed comparison across all twelve domain adaptation task in the small Amazon benchmark data. The reviews are from the domains *Books, Kitchen appliances, Electronics, DVDs*. With an exception of $B \to E$ and $D \to E$, mSDA$_5$ leads to the lowest transfer-loss.

Each mapping $\mathbf{W}^s$ can be solved in closed-form as in (5), following the method described in section 3.2. We define the output of the first layer in the resulting mSDA as the average of all reconstructions,

$$\mathbf{h}^1 = \tanh\left(\frac{1}{S}\sum_{s=1}^S \mathbf{W}^s \mathbf{x}^s\right). \quad (9)$$

Once the first layer, of dimension $r \ll d$, is built, we can stack multiple layers on top of it using the regular mSDA as described in section 3.3 and Algorithm 2. It is worth pointing out that, although features might be separated in different sub-sets within the first layer, they can still be combined in subsequent layers of the mSDA.

## 5. Results

We evaluate mSDA on the *Amazon reviews* benchmark data sets (Blitzer et al., 2006) together with several other algorithms for representation learning and domain adaptation. The dataset contains more than 340,000 reviews from 25 different types of products from Amazon.com. For simplicity (and comparability), we follow the convention of (Chen et al., 2011b; Glorot et al., 2011) and only consider the binary classification problem whether a review is positive (higher than 3 stars) or negative (3 stars or lower). As mSDA and SDA focus on feature learning, we use the raw bag-of-words (bow) unigram/bigram features as their input. To be fair to other algorithms that we compare to, we also pre-process with tf-idf (Salton & Buckley, 1988) and use the transformed feature vectors as their input if that leads to better results. Finally, we remove five domains which contain less than 1,000 reviews.

Different domains in the complete set vary substantially in terms of number of instances and class distribution. Some domains (books and music) have hundreds of thousands of reviews, while others (food and outdoor) have only a few hundred. There are a total of 380 possible transfer tasks (*e.g.* $Apparel \to Baby$). The proportion of negative examples in different domains also differs greatly. To counter the effect of class- and size-imbalance, a more controlled smaller dataset was created by Blitzer et al. (2006), which contains reviews of four types of products: books, DVDs, electronics, and kitchen appliances. Here, each domain consists of 2,000 labeled inputs and approximately 4,000 unlabeled ones (varying slightly between domains) and the two classes are exactly balanced. Almost all prior work provides results only on this smaller set with its more manageable *twelve* transfer tasks. We focus most of our comparative analysis on this smaller set but also provide results on the entire data for completeness.

**Methods.** As *baseline*, we train a linear SVM on the raw bag-of-words representation of the labeled *source* and test it on *target*. We also include the results of the same setup with dense features obtained by projecting the entire data set (labeled and unlabeled *source+target*) onto a low-dimensional sub-space with PCA (we refer to this setting as *PCA*). Besides these two baselines, we evaluate the efficacy of a linear SVM trained on features learned by mSDA and two alternative feature learning algorithms, Structural Correspondence Learning (*SCL*) (Blitzer et al., 2006) and 1-layer[1] *SDA* (Glorot et al., 2011). Finally, we also compare against *CODA* (Chen et al., 2011b), a state-of-the-art domain adaptation algorithm which is based on sample- and feature-selection, applied to tf-idf features. For CODA, SDA and SCL we use implementations provided by the authors. All hyper-parameters are set by 5-fold cross validation on the source training set[2].

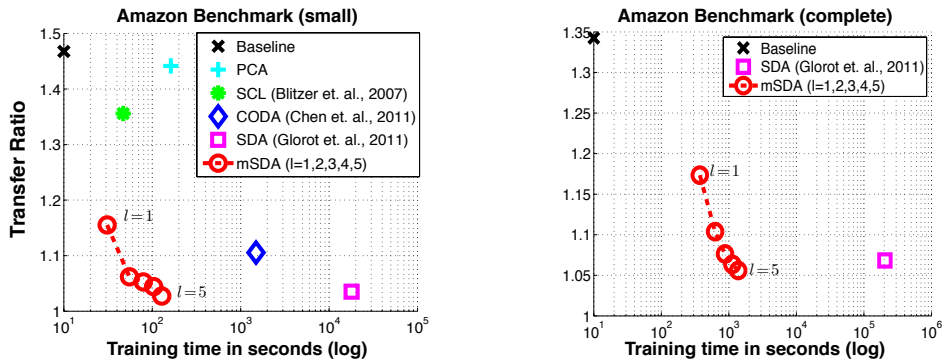

*Figure 2.* Transfer ratio and training times on the small (*left*) and full (*right*) Amazon Benchmark data. Results are averaged across the twelve and 380 domain adaptation tasks in the respective data sets ($5,000$ features). The graphs compare the results of mSDA with baseline, SDA and, on the small data set, with CODA, SCL and PCA. The speedups of mSDA over SDA, with similar transfer ratio, is $180\times$ on the small task and $230\times$ on the complete benchmark.

**Metrics.** Following Glorot et al. (2011), we evaluate our results with the *transfer error* $e(S, T)$ and the *in-domain error* $e(T, T)$. The *transfer error* $e(S, T)$ denotes the classification error of a classifier trained on the labeled *source* data and tested on the unlabeled *target* data. The *in-domain error* $e(T, T)$ denotes the classification error of a classifier that is trained on the labeled *target* data and tested on the unlabeled *target* data. Similar to Glorot et al. (2011) we measure the performance of a domain adaptation algorithm in terms of the *transfer loss*, defined as $e(S, T) - e_b(T, T)$, where $e_b(T, T)$ defines the in-domain error of the baseline. In other words, the transfer loss measures how much higher the error of an *adapted* classifier is in comparison to a linear SVM that is trained on actual *labeled target* bow data.

The various domain-adaptation tasks vary substantially in difficulty, which is why we do not average the transfer losses (which would be dominated by a few most difficult tasks). Instead, we average the *transfer ratio*, $e(S, T)/e_b(T, T)$, the ratio of the *transfer error* over the *in-domain error*. As with the *transfer loss*, a lower *transfer ratio* implies better domain adaptation.

---
[1]We were only able to obtain the 1-layer implementation from the authors. Anecdotally, multiple-layer *SDA* only leads to small improvements on this benchmark set but increases the training time drastically.

[2]We keep the default values of some of the parameters in SCL, *e.g.* the number of stop-words removed and stemming parameters — as they were already tuned for this benchmark set by the authors.

For timing purposes, we ignore the time of the SVM training and only report the mSDA or SDA training time. As both algorithms are unsupervised, we do not re-train for different transfer tasks within a benchmark set — instead we learn one representation on the union of all domains. CODA (Chen et al., 2011a) does not take advantage of data besides source and target and we report the average training time per transfer task.[3] All experiments were conducted on an off-the-shelf desktop with dual 6-core Intel i7 CPUs clocked at 2.66Ghz.

### 5.1. Comparison with Related Work

In the first set of experiments, we use the setting from (Glorot et al., 2011) on the small Amazon benchmark set. The input data is reduced to only the $5,000$ most frequent terms of unigrams and bigrams as features.

**Comparison per task.** Figure 1 presents a detailed comparison of the transfer loss across the twelve domain adaptation tasks using the various methods mentioned. A linear SVM trained on the features generated by SDA and mSDA clearly outperform all the other methods. For several tasks, the transfer loss goes to negative — in other words, a SVM trained on the transformed *source* data has higher accuracy than one trained on the original *target* data. This is a strong indication that the learned new representation bridges the gap between domains. It is worth pointing out that in ten out of the twelve tasks mSDA achieves a lower transfer-loss than SDA.

**Timing.** Figure 2 (left) depicts the transfer ratio as a function of training time required for different algorithms, averaged over 12 tasks. The time is plotted in log scale. We can make three observations: 1. SDA outperforms all other related work in terms of transfer-ratio, but is also the slow-est to train (more than 5 hours of training time). 2. SCL and PCA are relatively fast, but their features cannot compete in terms of transfer performance. 3. The training time of mSDA is two orders of magnitude faster that of SDAs (180× speedup), with comparable transfer ratio. Training one layer of mDA on all 27, 677 documents from the small set requires less than 25 seconds. A 5-layer mSDA requires less than 2 minutes to train, and the resulting feature transformation achieves slightly better transfer ratio than SDAs.

---
[3]In CODA, the feature splitting and classifier training are inseparable and we necessarily include both in our timing.

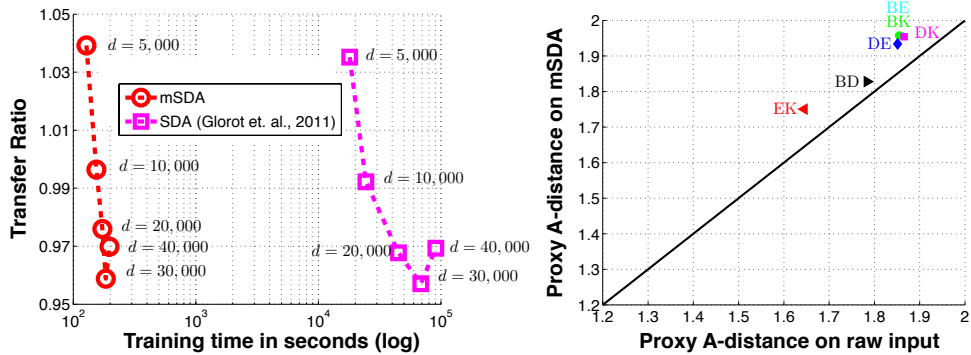

*Figure 3. Left:* Transfer ratio as a function of the input dimensionality (terms are picked in decreasing order of their frequency). *Right:* Besides domain adaptation, mSDA *also* helps in domain *recognition* tasks.

**Large scale results.** To demonstrate the capabilities of mSDA to scale to large data sets, we also evaluate it on the complete set with $n = 340,000$ reviews from 20 domains and a total of 380 domain adaptation tasks (see right plot in Figure 2). We compare mSDA to SDA (1-layer). The large set is more heterogenous in terms of the number of domains, domain size and class distribution than the small set and both the transfer error and transfer ratio are averaged across 380 tasks. Nonetheless, a similar trend can be observed. The transfer ratio reported in Figure 2 (right) corresponds to averaged transfer errors of (*baseline*) 13.93%, ( *one-layer SDA*) 10.50%, (*mSDA, l = 1*) 11.50%, (*mSDA, l = 3*) 10.47%, (*mSDA, l = 5*) 10.33%. With only one layer, mSDA performs a little worse than *SDA* but reduces the training time from over two days to about five minutes (700× speedup). With three layers, mSDA matches the transfer-error and transfer-ratio of SDA and still only requires 14 minutes of training time (230× speedup).

### 5.2. Further Analysis

In addition to comparison with prior work, we also analyze various other aspects of mSDA.

**Low-frequency features.** Prior work often limits the input data to the most frequent features (Glorot et al., 2011). We use the modification from section 4 to scale mSDA (5-layers) up to high dimensions and include less-frequent uni-grams and bi-grams in the input (small Amazon set). In the case of SDA we make the first layer a dimensionality reducing transformation from $d$ dimensions to 5000.

The left plot in Figure 3 shows the performance of mSDA and SDA as the input dimensionality increases (words are picked in decreasing order of their frequency). The transfer ratio is computed relative to the baseline with $d = 5000$ feature. Clearly, both algorithms benefit from having more features up to 30, 000. mSDA matches the transfer-ratio of SDA consistently and, as the dimensionality increases, gains even higher speed-up. With 30, 000 input features, SDA requires over one day and mSDA only 3 minutes (458× speedup).

**Transfer distance.** Ben-David et al. (2007) suggest the Proxy-A-distance (PAD) as a measure of how different two domains are from each other. The metric is defined as $2(1 - 2\epsilon)$, where $\epsilon$ is the generalization error of a classifier (a linear SVM in our case) trained on the binary classification problem to distinguish inputs *between* the two domains. The right plot in Figure 3 shows the PAD before and after mSDA is applied. Surprisingly, the distance *increases* in the new representation — *i.e.* distinguishing between two domains becomes *easier* with the mSDA features. We explain this effect through the fact that mSDA is unsupervised and learns a generally better representation for the input data. This helps both tasks, distinguishing between domains and sentiment analysis (*e.g.* in the electronic-domain mSDA might interpolate the feature "dvd player" from "blue ray", both are not particularly relevant for sentiment analysis but might help distinguish the review from the *book* domain.). Glorot et al. (2011) observe a similar effect with the representations learned with SDA.

### 5.3. General Trends

In summary, we observe a few general trends across all experiments: 1. With one layer, mSDA is up to three orders of magnitudes faster but slightly less expressive than the original SDA. This can be attributed to the fact that mSDA has no hidden layer. 2. There is a clear trend that additional "deep" layers improve the results significantly (here, up to five layers). With additional layers, the mSDA fea-tures reach (and surpass) the accuracy of 1-layer SDA and still obtain a several hundred-fold speedup. 3. The mSDA features help diverse classification tasks, domain classification and sentiment analysis, and can be trained very efficiently on high-dimensional data.

## 6. Discussion and Conclusion

Although mSDA first and foremost marginalizes out the corruption in SDA training, the two algorithms differ in several profound ways: First, the mDA layers do not have hidden nodes — this allows a closed-form solution with substantial speed-ups but might entail limitations that still need to be investigated. Second, mSDA only has *two* free meta-parameters, controlling the amount of noise as well as the number of layers to be stacked, which greatly simplifies the model selection. Finally, leveraging on the analytic tractability of linear regression, the parameters of an mDA are trained to optimally denoise *all possible* corrupted training inputs — arguably "*infinitely many*". This is practically infeasible for SDAs.

We hope that our work on mSDA will inspire future research on efficient training of SDA, beyond domain adaptation, and impact a variety of research problems. The fast training time, the capability to scale to large and high-dimensional data and implementation simplicity make mSDA a promising method with appeal to a large audience within and beyond machine learning.

## Acknowledgements

KQW, MC, ZX were supported by NSF IIS-1149882 and NIH U01 1U01NS073457-01. FS was supported by NSF IIS-0957742, DARPA CSSG N10AP20019 and D11AP00278.